 \documentclass[conference]{IEEEtran}
\IEEEoverridecommandlockouts

\usepackage{graphicx}
\usepackage{subcaption}
\usepackage[linesnumbered,ruled,vlined]{algorithm2e}
\usepackage{color}
\usepackage{url}
\usepackage{amsmath}
\usepackage{algpseudocode}
\usepackage{booktabs}
\usepackage{xcolor}
\usepackage{flushend}

\usepackage[pass,top=0.73in]{geometry}
\title{GCN-Based Throughput-Oriented Handover Management in Dense 5G Vehicular Networks

\thanks{This work is partially funded by NSERC Discovery grants.}
}

\author{\IEEEauthorblockN{Nazanin Mehregan}
    \IEEEauthorblockA{Department of Computer Science \\ Brock University, Canada \\
   nmehregan@brocku.ca}
   
    \and
    \IEEEauthorblockN{Robson E. De Grande}
    \IEEEauthorblockA{Department of Computer Science \\ Brock University, Canada \\
    rdegrande@brocku.ca}

}

\begin{document}

\maketitle

\begin{abstract}
The rapid advancement of 5G has transformed vehicular networks, offering high bandwidth, low latency, and fast data rates essential for real-time applications in smart cities and vehicles. These improvements enhance traffic safety and entertainment services. However, 5G’s limited coverage and frequent handovers, causing network instability from the "ping-pong effect," pose challenges in high-mobility environments. This paper presents TH-GCN (Throughput-oriented Graph Convolutional Network), a novel approach for optimizing handover management in dense 5G networks. Utilizing graph neural networks (GNNs), TH-GCN models vehicles and base stations as nodes in a dynamic graph with enriched features like signal quality, throughput, vehicle speed, and base station load. Integrating both user equipment and base station perspectives, this dual-centric approach enables adaptive, real-time handover decisions that improve stability. Simulations show that TH-GCN reduces handovers by up to 78\% and improves signal quality by 10\%, outperforming existing methods and positioning it as a key advancement in 5G vehicular networks.

\end{abstract}


\section{Introduction}
Vehicular Networks (VNs) are essential to Intelligent Transportation Systems (ITS), enabling real-time applications that enhance traffic safety, efficiency, and in-vehicle entertainment, though establishing reliable, high-bandwidth, low-latency connections in urban settings remains challenging~\cite{8376311}. Vehicle-to-Everything (V2X) communication is central to this advancement, facilitating continuous interactions between vehicles and network elements. The introduction of fifth-generation (5G) mobile networks, particularly with millimetre-wave (mmWave) technology, offers unprecedented solutions to meet these high data demands, enabling advanced connectivity in high-mobility scenarios such as vehicular networks~\cite{choi2016millimeter}. The high-speed, low-latency capabilities of 5G enable advanced applications like real-time communication in smart vehicles and smart cities, and its high bandwidth supports data-intensive tasks such as multimedia streaming~\cite{lai2020security}. However, the high-frequency nature of mmWave presents challenges; signals are easily obstructed by urban obstacles, resulting in intermittent connectivity and frequent disruptions. Additionally, as vehicle density increases, network quality declines, and the limited coverage range of 5G towers requires a high-density deployment to maintain connectivity~\cite{kosmopoulos2022handover}. This dense setup, paired with high vehicle mobility, increases the "ping-pong effect," where vehicles rapidly switch between neighbouring towers. Such frequent handovers compromise stability and quality of service (QoS), particularly in urban areas~\cite{hasan2018frequent}.

Handover management in 5G networks presents critical challenges, particularly in high-mobility vehicular environments. The dense deployment of small cells in 5G networks—while essential for achieving ultra-reliable low-latency communication (URLLC)—exacerbates issues such as frequent handovers (HOs), HO failures, and ping-pong effects, which degrade throughput and increase latency~\cite{haghrah2023survey}. For instance, studies on 5G-NR networks reveal that ultra-dense cell deployments lead to an increase in HO frequency compared to 4G networks, directly impacting QoS for vehicular users~\cite{bilen2017handover}. Furthermore, the dynamic nature of vehicular networks amplifies these challenges: vehicles moving at 60–120 km/h require sub-second HO decisions to avoid radio link failures (RLFs), a problem inadequately addressed by traditional HO algorithms~\cite{hong2021modified}.

To address handover management challenges in 5G Vehicular Networks, we propose TH-GCN (Throughput-oriented Graph Convolutional Network), an intelligent solution that optimizes real-time handover decisions. Traditional handover management approaches often treat handover decisions as isolated events, failing to capture the broader network dynamics. TH-GCN addresses this limitation by introducing a graph-oriented modelling approach that considers spatial and temporal dependencies explicitly. The system models the network as a dynamic graph where vehicles and base stations serve as nodes with specific features (speed, direction, and load), while edges encode multi-parameter relationships (signal quality, throughput, and distance). This approach effectively captures spatial and temporal dependencies that conventional reinforcement learning (RL) and deep learning (DL) methods typically overlook~\cite{huang2020efficient, liu2021spatio}. TH-GCN reduces unnecessary handovers and ping-pong effects by 75\% compared to RL-based methods while enabling real-time inference on edge devices through lightweight graph convolutions~\cite{mubashir2023}. Unlike existing predictive models that require GPU clusters for training, our approach's computational efficiency makes it suitable for edge deployment. The system's topology awareness distinguishes it from prior works that rely on static network models, as its dynamic graph structure continuously adapts to real-time changes in vehicle trajectories and handovers, leading to improved handover accuracy.

Furthermore, TH-GCN implements multi-stakeholder optimization, advancing beyond traditional methods that focus solely on user equipment (UE) metrics like signal strength by jointly optimizing for base station load and network-wide throughput. Using spatiotemporal graph embeddings, the system predicts future handover targets. While existing predictive handover frameworks face limitations—RL-based approaches struggle with scalability in ultra-dense networks due to centralized control, and DL-based methods fail to generalize across heterogeneous network topologies—TH-GCN addresses these challenges by unifying predictive accuracy with interpretable graph structures. Simulations in 5G New Radio (5G-NR) environments validate our approach as a robust, scalable, and efficient solution for next-generation vehicular networks.

The paper is organized as follows. Section~\ref{sec:related_work} reviews handover management strategies.
Section~\ref{sec:methodology} introduces the TH-GCN framework.
Section~\ref{sec:pa} presents a quantitative analysis of TH-GCN’s performance and results. Finally, Section~\ref{sec:conclusion_future_works} summarizes key findings and suggests future research directions.

\section{Related Work}
\label{sec:related_work}

Handover management in 5G networks has been a focal point of research, particularly with the advent of ultra‐dense networks (UDNs) and the increasing complexity of vehicular environments. Traditional methods, such as reactive and proactive handover strategies, have been widely studied; however, they often fail to address the dynamic, heterogeneous, and interconnected nature of modern 5G deployments.~\cite{andrews2014will}. In vehicular environments—where high mobility and frequent topology changes further complicate network operations—classical algorithms may lead to excessive handover overhead and instability~\cite{xenakis2013mobility}. Recent advancements in machine learning, particularly deep reinforcement learning (DRL) have shown promise in improving handover management by learning to predict optimal handover timings and target cells~\cite{khosravi2020learning}. Yet, these methods still face significant challenges in capturing complex spatial dependencies and adapting to non-linear network dynamics. To address these issues, our proposed TH-GCN (Throughput-oriented Graph Convolutional Network) leverages graph neural networks (GNNs) to model inter-cell relationships, enabling more accurate and efficient handover parameter optimization~\cite{mehrabi2023neighbor, foka2023handover}.

Efficient handover management is essential for maintaining seamless connectivity in wireless communication systems, enabling smooth transfers of active communication sessions as users move. Reactive, or event-triggered, handover methods initiate handovers only when specific conditions, such as signal strength, are met~\cite{lopez2012mobility}. This approach minimizes unnecessary transitions and reduces signalling overhead. For example, Lopez et al.~\cite{lopez2012mobility} implemented a Time-to-Trigger (TTT) parameter to delay handovers until network conditions are stable for a set duration. Qi et al.~\cite{qi2020federated} enhanced these methods for mmWave vehicular networks by incorporating signal-to-noise ratio (SNR) thresholds, effectively reducing handover frequency and mitigating the ping-pong effect. Furthermore, optimizing metrics like Carrier to Interference-and-Noise Ratio (CINR) and Received Signal Strength Indicator (RSSI) can further minimize unnecessary handovers.

In contrast, proactive handover methods predict future network conditions to initiate handovers before signal deterioration occurs. These techniques leverage mobility and channel predictions, drawing from historical data to optimize decisions~\cite{wang2018handover, alkhateeb2018machine, lee2020prediction, sun2017smart}. Chih-Lin I. et al.~\cite{chih2017big} employ support vector machines to reduce service interruptions in LTE networks, while~\cite{wang2018handover} applies deep reinforcement learning to minimize frequent handovers in ultra-dense environments. Proactive strategies are particularly effective in addressing sudden signal attenuation in mmWave networks~\cite{sun2017smart, alkhateeb2018machine, lee2020prediction}. For instance, Alkhateeb et al.~\cite{alkhateeb2018machine} proposed a deep-learning model that predicts line-of-sight blockages in mmWave networks, achieving 95\% accuracy and improving network reliability. Qi et al.~\cite{qi2020federated} also introduced a federated learning framework that enables decentralized, privacy-preserving training on local data, adapting to changes in user mobility. Additionally, Sahin et al.~\cite{sahin2018virtual} presented a hybrid user-centric radio access virtualization strategy to optimize vehicle-to-everything (V2X) communications in 5G networks.

Threshold-based methods optimize handover decisions through multi-attribute metrics. For example, Sun et al.~\cite{sun2021multi} proposed a link-reliability-throughput (LRT) strategy that adapts to dense urban environments, improving network performance by balancing Quality of Service (QoS) across metrics. Reinforcement learning (RL) techniques have also been applied to optimize handover decisions adaptively, such as the SMART policy proposed by Sun et al.~\cite{sun2017smart}, which uses deep neural networks to learn optimal handover timings based on channel conditions and user requirements. In vehicular networks, Mubashir et al.~\cite{mubashir2023} introduced a SARSA RL algorithm, dynamically adjusting handover decisions by evaluating parameters such as signal strength, tower load, and user speed. This approach enhances connection stability by minimizing the number of handovers and adjusting to real-time network conditions. These proactive methods, however, often require extensive computational resources and may not fully capture the intricate spatial relationships among base stations and user equipment in ultra-dense networks.

A recent trend has been the application of Graph Neural Networks (GNNs) to address the limitations of traditional ML approaches. GNNs are naturally suited to model the complex interdependencies in a network, as they operate on graph-structured data. Early studies, such as those by Zhao et al.~\cite{zhao2019t} and Li et al.~\cite{li2021detectornet}, primarily applied GNNs for traffic forecasting. For example, T-GCN~\cite{zhao2019t} combines graph convolutional networks (GCNs) with gated recurrent units (GRU) to capture spatio-temporal correlations for urban traffic prediction. Similarly, DetectorNet~\cite{li2021detectornet} leverages graph-based representations to enhance traffic prediction accuracy. Although these models excel in capturing spatial and temporal dependencies, their focus is on forecasting traffic patterns rather than directly optimizing handover decisions. In other words, while they successfully predict network congestion or traffic flow, they do not directly tackle the problem of selecting the best tower for handovers, nor do they explicitly consider throughput optimization, which directly measures and supports service quality in upper layers. 

More recent GNN-based approaches in the handover domain have been presented by Mehrabi et al.~\cite{mehrabi2023neighbor} and Djuikom Foka et al.~\cite{foka2023handover}. Mehrabi et al. propose an auto-grouping GCN framework for handover parameter configuration that clusters neighbouring cells to optimize local performance. Djuikom Foka et al. introduced a spatiotemporal GNN to forecast handover events by modelling the network as a graph where nodes represent base station (BS) pairs. Although these works mark important steps toward incorporating GNNs in handover management, they typically focus on either parameter configuration or forecasting the number of handovers rather than optimizing the quality of the connection as measured by throughput. Similarly, Ji et al.~\cite{ji2024graph} integrate GNNs with DRL for resource allocation in V2X communications, optimizing spectrum and power management, but their approach focuses on resource efficiency rather than handover optimization. In contrast, TH-GCN specifically addresses handover management by optimizing throughput and minimizing unnecessary handovers, providing a more holistic solution for network performance in 5G vehicular environments.


Traditional reactive and proactive handover methods in 5G networks, while advancing the field, struggle to capture the complex spatial dependencies and dynamic throughput requirements of ultra-dense, high-mobility environments, and even recent GNN-based approaches tend to focus on traffic forecasting or parameter configuration rather than optimizing network performance. Our proposed TH-GCN overcomes these shortcomings by formulating handover management as a multi-objective optimization problem that integrates throughput optimization into its graph convolutional framework through edge-weighted aggregation, triplet loss optimization, and a dynamic graph structure. This comprehensive solution minimizes unnecessary handovers and the ping-pong effect while maximizing overall network throughput, offering a robust, scalable, and computationally efficient approach for next-generation 5G vehicular networks.

\section{TH-GCN Model}
\label{sec:methodology}

The proposed TH-GCN (Throughput-Oriented Graph Convolutional Network) model is designed to enhance handover management in 5G vehicular networks, tackling issues like high vehicle mobility, limited coverage range, and frequent ping-pong handovers. The primary objective of TH-GCN is to reduce handover occurrences while maximizing both throughput and signal quality in ultra-dense urban 5G settings. This optimization problem is framed as a multi-objective function with constraints that limit the total number of handovers, reduce tower load variance, and maintain minimum throughput and signal quality. The model seeks the best trade-offs among these objectives within the defined constraints.

\begin{figure*}[t!]
  \centering
  \includegraphics[width=0.92\textwidth]{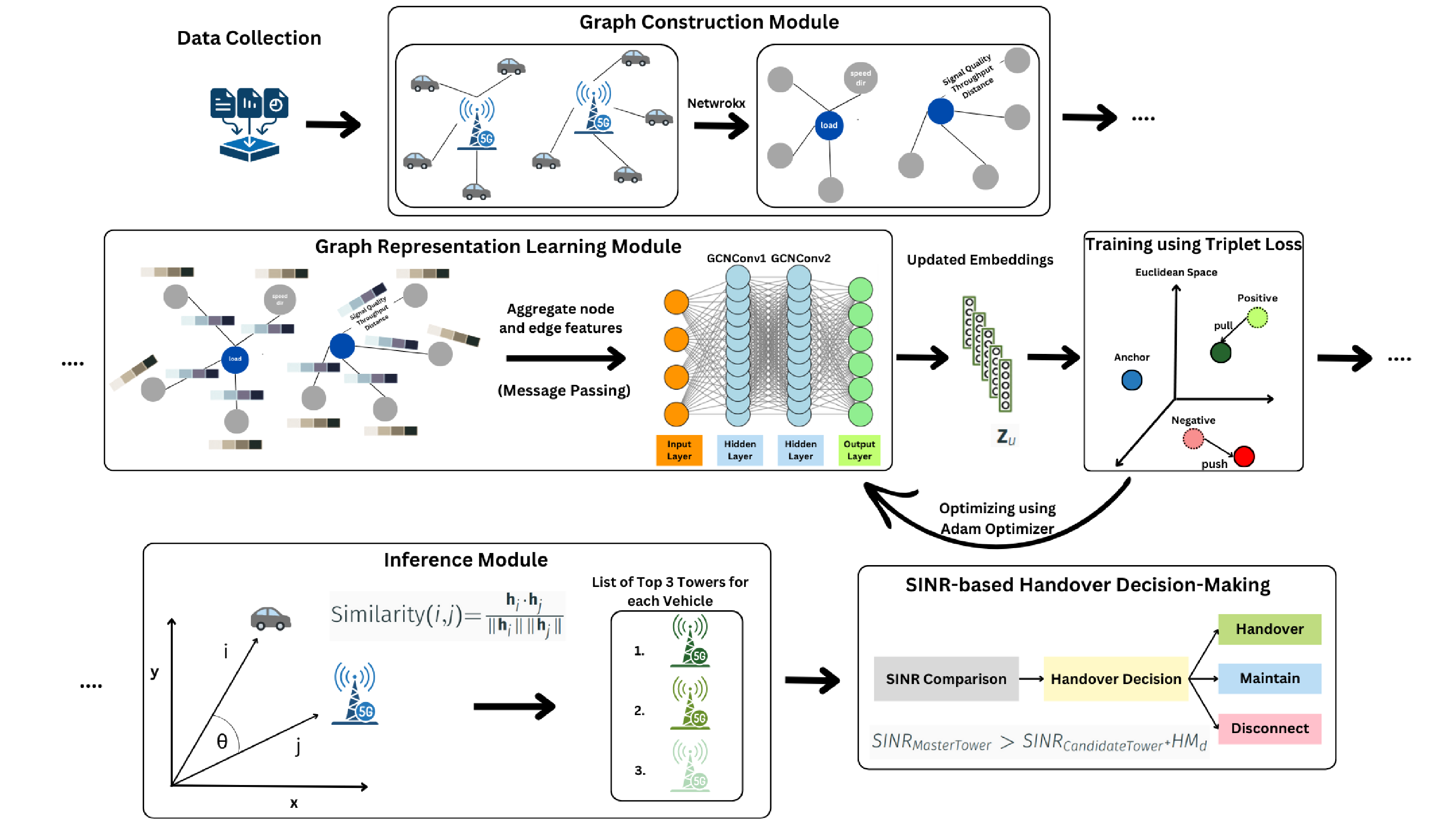} 
  \vspace{0mm}
  \caption{Overview of the TH-GCN pipeline for optimizing handover decisions in dense 5G vehicular networks.}
  \label{fig:pipeline}
\end{figure*}

Our approach to solving this problem leverages a dynamic graph structure that updates at each timestamp to capture node features and edge interactions, creating a data-driven, real-time solution for effective tower selection, as depicted in Figure~\ref{fig:pipeline}.



\subsection{Dataset Collection}

Our dataset is generated from a simulated 5G vehicular network using Omnetpp and Simu5G, capturing key interactions between vehicles and towers that impact handover decisions. These interactions are structured as a graph, with specific features mapped onto nodes (vehicles and towers) and edges (vehicle-tower connections) for effective analysis within a GNN framework.


The dataset includes several features essential for understanding network behaviour and predicting optimal handover events:

\begin{itemize}
    \item \textbf{Throughput:} Represents the data rate successfully delivered to a vehicle, calculated as:
    \begin{equation}
        \footnotesize
        \text{Throughput} = \frac{\text{Total data sent (in bytes)}}{\text{Transmission time (in seconds)}}
    \end{equation}
    \item \textbf{Signal Quality:} Quantified by Signal-to-Interference-Plus-Noise Ratio (SINR), indicating the communication link’s reliability.
    \item \textbf{Mobility:} Captures the vehicle's speed and direction, providing insights into connectivity changes and handover needs.
    \item \textbf{Tower Load:} Reflects the current number of vehicles connected to each tower, affecting tower capacity.
    \item \textbf{Positions:} Geographic coordinates of vehicles and towers, providing spatial context for network interactions.
\end{itemize}

\subsection{Graph Construction and Structure}

Each simulation timestamp is represented as a graph where vehicles and towers are nodes, and the connections between them form edges. This dynamic graph includes heterogeneous nodes (vehicles and towers) and evolving connections, capturing real-time mobility and network conditions.

\subsubsection{Node and Edge Features}

\begin{itemize}
    \item \textbf{Node Features:} Vehicle nodes are initialized with a feature vector comprising features such as speed, direction, and position, while tower nodes include attributes like load.
    \item \textbf{Edge Features:} Connections between vehicles and towers are weighted by metrics, such as throughput, signal quality, and distance.
\end{itemize}

\subsubsection{Graph Representation}

Graphs are represented using NetworkX for visualization and PyTorch Geometric for efficient processing. This enables the GNN model to incorporate dynamic changes in network structure and interaction strengths.

\subsection{Graph Neural Network (GNN) Architecture}

Our GNN-based model employs a Graph Convolutional Network (GCN) to capture the relationships in vehicular networks. The architecture includes message passing, embedding generation, and a customized loss function to optimize tower selection. 
GCNs efficiently aggregate information from neighbouring nodes in small to medium-sized graphs, reducing computation and memory needs. This enables quick message passing, making them ideal for real-time applications in vehicular networks. Their interpretability also helps in understanding the model's decision-making for optimizing handover decisions.

\subsubsection{Message Passing and Node Embedding}

The GCN aggregates node and edge features through iterative message-passing steps, updating each node's representation to incorporate the influence of its neighbouring nodes and edges. The process can be formalized as:
\begin{equation}
    \scriptsize
    h_{u}^{(k+1)} = \text{UPDATE}^{(k)} \left( h_{u}^{(k)}, \text{AGGREGATE}^{(k)} \left( \{h_{v}^{(k)} \mid \forall v \in N(u)\} \right) \right)
\end{equation}
where \(h_{u}^{(k+1)}\) denotes the updated feature vector for node \(u\) at layer \(k+1\), and \(N(u)\) represents the neighbors of \(u\). After aggregation, the aggregated message is passed through a dense neural network layer (a GCN convolution layer in this case), which is a transformation using a weight matrix and a non-linear activation function (like ReLU). This gives the updated node features for each node at layer k+1.

\subsubsection{Embedding Space and Optimization}

The GCN generates embeddings in a lower-dimensional space, positioning nodes with similar connectivity roles and attributes closer together. After two layers of GCN—each performing a step of message passing, aggregation, and feature update—the final node embeddings are obtained at the last layer. These lower-dimensional vectors effectively encode the structural and feature information of the nodes within their graph neighbourhood, capturing both the individual attributes of each node and the influence of its connected nodes. This embedding space facilitates effective handover prediction by encoding each vehicle’s connectivity quality and tower suitability.

\subsection{Model Training}

The TH-GCN model is trained using an incremental, unsupervised learning approach that leverages a triplet loss function to optimize node embeddings for effective tower selection. Dynamic and heterogeneous graphs are initially constructed from the dataset, capturing the interactions between vehicles and towers at each simulation timestamp. The following subsections detail the key components of the training process.

\subsubsection{Initialization and Incremental Training}

The Graph Convolutional Network (GCN) parameters are initialized either by loading previously saved parameters or by random initialization if no prior training exists. This setup facilitates incremental training, where the optimized parameters are saved after each training iteration. These saved parameters are loaded in subsequent sessions, allowing the model to continue training without retraining from scratch. This incremental approach enhances efficiency and enables the model to adapt continuously to evolving network conditions.

\subsubsection{Triplet Loss Optimization}

To ensure that vehicles are positioned closer to positively connected towers than to negative, disconnected towers, the model uses a triplet loss function:
\begin{equation}
    \footnotesize
    \mathcal{L}(a, p, n) = \max \left( d(f(a), f(p)) - d(f(a), f(n)) + \alpha, 0 \right)
\end{equation}
where \(d(x, y)\) denotes a distance metric (typically Euclidean distance), \(a\) is the anchor (vehicle), \(p\) is the positive sample (connected tower), \(n\) is the negative sample (disconnected tower), and \(\alpha\) is the margin parameter. This loss function encourages the model to learn embeddings where vehicles are closer to their optimal towers and farther from suboptimal ones, enhancing tower ranking accuracy.

\subsubsection{Edge-Weighted Aggregation}

Unlike standard GCNs, which treat all edges equally, our model incorporates edge weights (e.g., throughput, signal quality, and distance) to prioritize high-quality connections. The modified update rule is:
\begin{equation}
    \footnotesize
    H^{(l+1)} = \sigma\left(\hat{D}^{-\frac{1}{2}} (\hat{A} \odot W) \hat{D}^{-\frac{1}{2}} H^{(l)} W^{(l)}\right)
\end{equation}
where \(W\) represents edge weights, \(\odot\) denotes element-wise multiplication, and \(\sigma\) is the activation function. This edge-weighted aggregation enables the model to focus on high-throughput connections, enhancing tower selection accuracy.

\subsubsection{Training Pipeline}

The training process includes the following steps:

\begin{enumerate}
    \item Graph construction, incorporating node and edge features.
    \item Initialization of model parameters and node embeddings via GCN layers.
    \item Iterative training with triplet loss to refine the embedding space for optimal tower selection.
\end{enumerate}

\subsection{Inference and Tower Ranking}

During the inference phase, the trained TH-GCN model generates node embeddings for each vehicle and tower, enabling a similarity-based ranking of potential towers. A similarity matrix is constructed to quantify the affinity between each vehicle and potential towers. The similarity score \( S_{v,t} \) for each vehicle \( v \) and each tower \( t \) is calculated using a similarity metric, typically cosine similarity:
\[
S_{v,t} = \frac{E_{final}^v \cdot E_{final}^t}{\| E_{final}^v \| \| E_{final}^t \|}
\]
where \( E_{final}^v \) and \( E_{final}^t \) represent the embeddings of vehicle \( v \) and tower \( t \), respectively. Here, \( S_{v,t} \) indicates how suitable a tower is as a candidate for handover based on the learned embeddings.

For each vehicle \( v_i \), towers \( u_j \) are ranked based on these similarity scores, ensuring that handover decisions are both efficient and Quality of Service (QoS) optimized. By selecting the top candidate towers based on similarity scores, the model reduces interruptions and maximizes throughput, ultimately improving network stability and user experience. The entire pipeline is summarized in Algorithm~\ref{algo:th_gcn}.

\begin{algorithm}[t]
    \caption{TH-GCN: Handover Management}
    \label{algo:th_gcn}
    \SetAlgoLined
    \KwIn{Dataset \(D\), Simulation timestamp \(t\)}
    \KwOut{Top 3 candidate towers \(\Omega_v\) for each vehicle \(v\)}
    
    Initialize GNN parameters \(\theta\)\;
    Extract graph \(G(t)\), node features \(X_{node}(t)\), and edge features \(X_{edge}(t)\) from dataset \(D\) for timestamp \(t\)\;
    
    \If{model parameters saved}{
        Load saved parameters\;
    }
    \Else{
        Initialize model\;
    }
    
    \While{training}{
        Compute node embeddings: \(E_{node}(t) \gets GNN(G(t), X_{node}(t); \theta_{gnn})\)\;
        
        \For{each vehicle \(v\) in \(G(t)\)}{
            Select connected tower \(t^+\) and non-connected tower \(t^-\)\;
            Append triplet \((E_{node}^v, E_{node}^{t^+}, E_{node}^{t^-})\) to set \(T\)\;
        }
        
        Compute triplet loss \(L \gets \text{TripletLoss}(T; \theta_{loss})\)\;
        Backpropagate loss and update \(\theta\) using gradient \(\nabla_\theta L\)\;
    }
    
    Save trained model parameters\;
    
    Compute final embeddings: \(E_{final} \gets GNN(G(t), X_{node}(t); \theta_{gnn})\)\;
    
    \For{each vehicle \(v\)}{
        \For{each tower \(t\)}{
            Compute similarity score \(S_{v,t} \gets \text{similarity}(E_{final}^v, E_{final}^t)\)\;
        }
    }
    
    \For{each vehicle \(v\)}{
        Select top 3 towers: \(\Omega_v \gets \text{Top3}(S_{v,:})\)\;
    }
    
    \KwRet{\(\Omega_v\) for all vehicles}
\end{algorithm}

\subsection{Handover Decision Execution}
The handover decision process involves evaluating candidate towers by comparing their Received Signal Strength Indicator (SINR) with the current tower’s SINR, factoring in a hysteresis threshold to prevent unnecessary handovers. A minimum SINR threshold is also used to ensure sufficient signal quality. Based on these criteria, the system either maintains the current connection, initiates a handover to a better tower, or disconnects if the signal quality is too low.




%
%
%
\section{Performance Analysis and Results}
\label{sec:pa}

The performance analysis was conducted in a comprehensive simulation environment combining INET 4.3, Simu5G 1.1.0, VEINS 5.1, SUMO, and OMNet++. INET provides core network protocols, with Simu5G enabling advanced 5G functionality and VEINS supporting realistic vehicular mobility modelling to study communication protocols in vehicular contexts. SUMO simulates large-scale traffic, all coordinated through OMNet++ 6.0 (pre10) to ensure integration across frameworks.

\begin{figure}[t!]
  \centering
  \includegraphics[width=.45\textwidth]{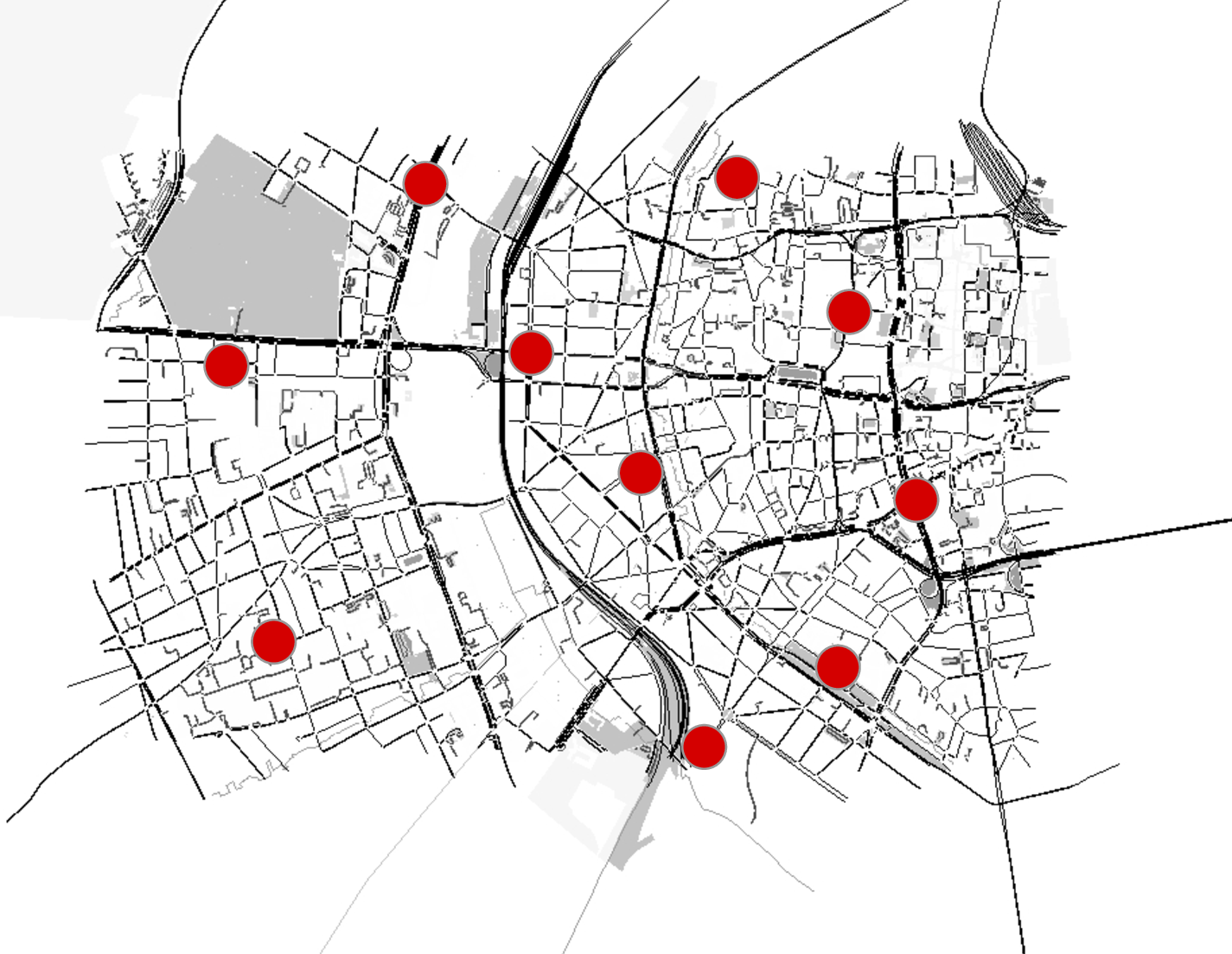}
  \vspace{-3mm}
    \caption{Simulation Scenario of Cologne, Germany.}
    \label{fig:CologneMap}
\end{figure}

The simulation is based on a real-world map of Cologne, Germany, shown in Figure~\ref{fig:CologneMap}, featuring ten strategically placed 5G gNodeB towers to create a realistic network environment with both overlapping and gap-filled coverage areas. Vehicles follow distinct routes between various start and destination points, observing traffic rules. Multiple runs with varied routes allow for an in-depth network performance analysis across different conditions.

\subsection{Network Application}
Our simulation framework uses a VoIP application in Simu5G to generate continuous uplink traffic from vehicles to 5G gNodeB towers, emulating real-world VoIP usage in vehicular networks. Packet IDs and timestamps are logged in voipSender, with arrival times recorded in voipReceiver, enabling precise, time-based throughput and packet loss calculations. We assess the average throughput for each vehicle-tower pair. This throughput data is then fed into our TH-GCN model to enhance handover management, supporting more accurate network performance analysis and optimized handover decisions. The specific parameters used in the VoIP application are detailed in Table \ref{table:simulationParameters}.

\subsection{Parameters Settings}

As shown in Table~\ref{table:simulationParameters}, the simulation parameter settings were configured to represent realistic urban traffic conditions. Vehicle speeds ranged from 0-180 km/h across a network of 10 gNodeB towers. We tested scenarios with varying vehicle densities (100-1000 vehicles in increments of 300) over 300-second simulations, with results averaged across 10 runs using different seeds at 95\% confidence intervals. The GNN model used 64 hidden channels over 50 epochs with a 0.01 learning rate, processing network data every 0.5 seconds.

\begin{table}[t]
  \centering
  \caption{Parameter Settings of the Performance Analyses.}
  \vspace{-2mm}
  \begin{tabular}{ll} 
    \toprule
    \textbf{Parameter} & \textbf{Value/Range}  \\  
    \midrule
    \multicolumn{2}{l}{\textbf{Simulation Environment}} \\
        Urban Area & 5000 x 5000 {$m^2$} \\
        Number of Vehicles & 100 - 1000  \\
        Vehicle Speed & 0 km/h - 180 km/h \\
        Number of Towers & 10 \\
        Comm. range (Tower) & 1000 m \\       
        Physical Model & 5G \\      
        gNodeB Trans. Power & 46 dBm \\
        Vehicle Trans. Power & 26 dBm \\
    \midrule
    \multicolumn{2}{l}{\textbf{GNN Model}} \\
        Learning Rate (lr) & 0.01 \\
        Triplet Loss Margin & 1.0 \\
        Hidden Channels & 64 \\
        Out Channels & 32 \\
        Epochs & 50 \\
    \midrule
    \multicolumn{2}{l}{\textbf{VoIP Application}} \\
        Configuration & VoIP Uplink \\
        VoipReceiver & gNodeB server \\
        VoIPSender & Vehicle \\
        Packet Size & 256 bytes \\
        Sampling Time & 0.02 s \\
        Bitrate & 102.4 kbps\\
    \midrule
    \multicolumn{2}{l}{\textbf{Handover Management}} \\
        SINR Sampling Time & 0.5 s \\
        GNN Script Execution & Every 0.5 - 5 - 10 s\\
    \bottomrule
  \end{tabular}
  \label{table:simulationParameters}
\end{table}

\subsection{Performance Metrics}

The following metrics were observed to measure the performance of our proposed approach.
\textbf{Average SINR} is the mean signal quality (signal-to-interference-plus-noise ratio) between vehicles and towers, reflecting the communication channel quality. 
\textbf{Average Simulation Throughput} is the total data successfully transferred between the vehicle and the tower during simulation, measured in bps, indicating the connection quality and network capacity for data applications like VoIP.
\textbf{Packet Transmission Rate} is the average rate at which packets are transmitted and received, focusing only on successfully delivered packets and excluding lost packets. This metric is crucial for assessing data efficiency on a per-packet basis, especially in real-time applications.
\textbf{Packet Loss Ratio and Packet Delivery Ratio} are the proportions of lost and successfully delivered packets, respectively, serving as indicators of data reliability and network quality.
\textbf{Handovers and Ping-pong Handovers} are the total number of base station transitions and rapid back-and-forth connections, respectively, which reflect network stability and handover efficiency.

\subsection{Results}

To assess the effectiveness of the proposed TH-GCN approach, we conducted comparative performance evaluations with two other handover decision algorithms: the Baseline~\cite{nardini2020simu5g} and CO-SRL~\cite{mubashir2023}. Key metrics include average SINR, throughput, packet transmission rate, packet loss ratio, total handovers, and ping-pong handovers. These metrics were measured across varying vehicle densities ($V = 100, 400, 700, 1000$) to simulate diverse network conditions.

\begin{figure*}[htp]
    \centering

    \begin{subfigure}{.45\textwidth}
        \centering 
        \includegraphics[trim={8pt 8pt 7pt 7pt},clip,width=\textwidth]{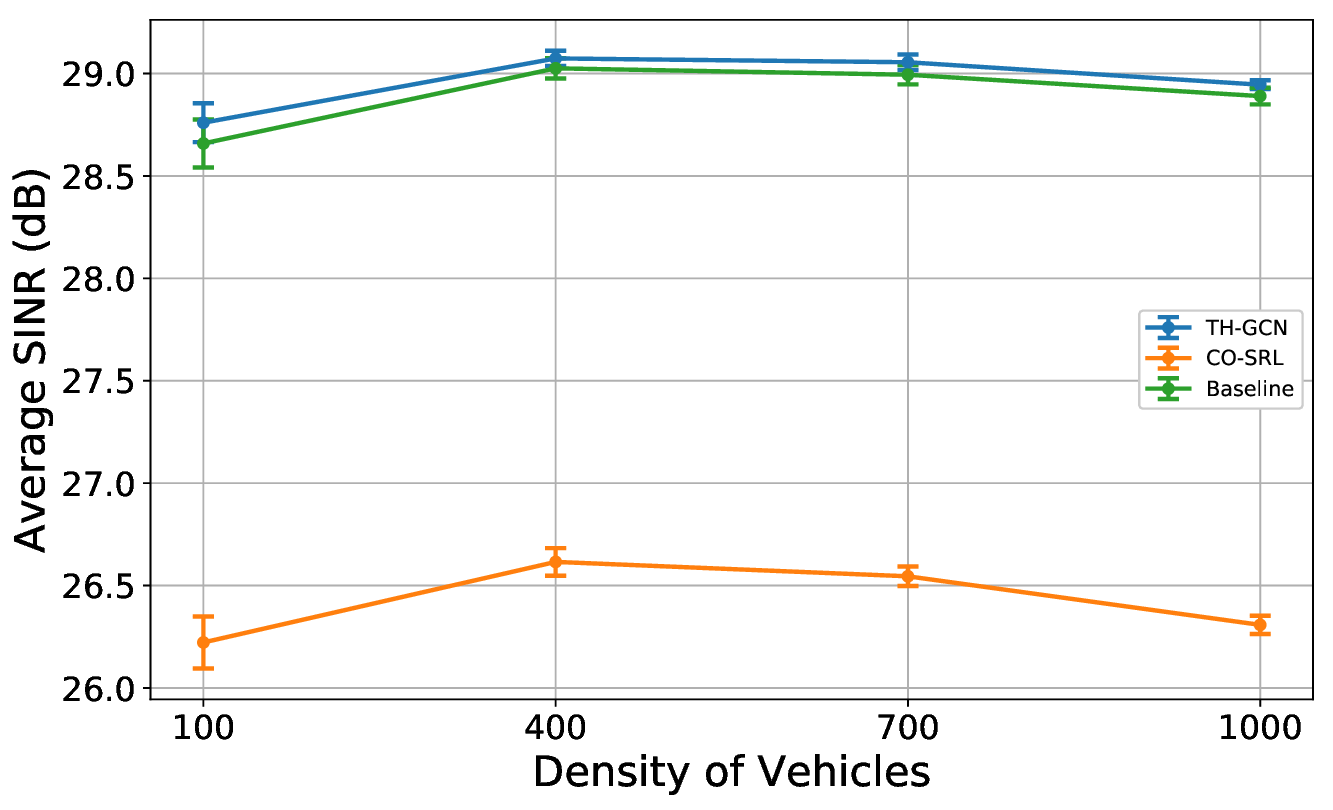}
        \vspace{-4mm}
        \caption{Average SINR}
        \label{fig:sinr}
    \end{subfigure}
    \;
    \begin{subfigure}{.45\textwidth}
        \centering
        \includegraphics[trim={8pt 5pt 7pt 7pt},clip,width=\textwidth]{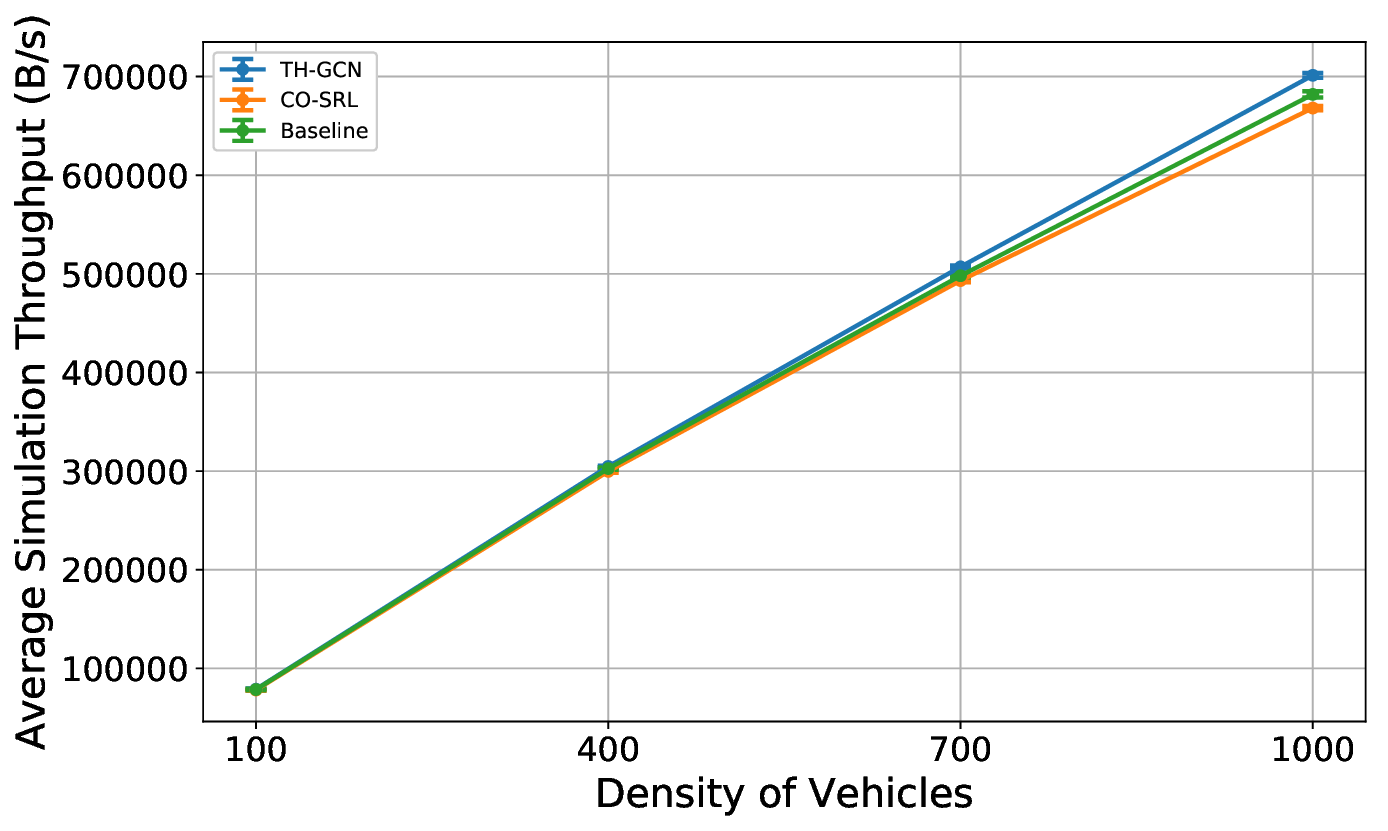}
        \vspace{-4mm}
        \caption{Average Throughput}
        \label{fig:throughput}
    \end{subfigure}

    \begin{subfigure}{.45\textwidth}
        \centering
        \includegraphics[trim={8pt 5pt 7pt 7pt},clip,width=\textwidth]{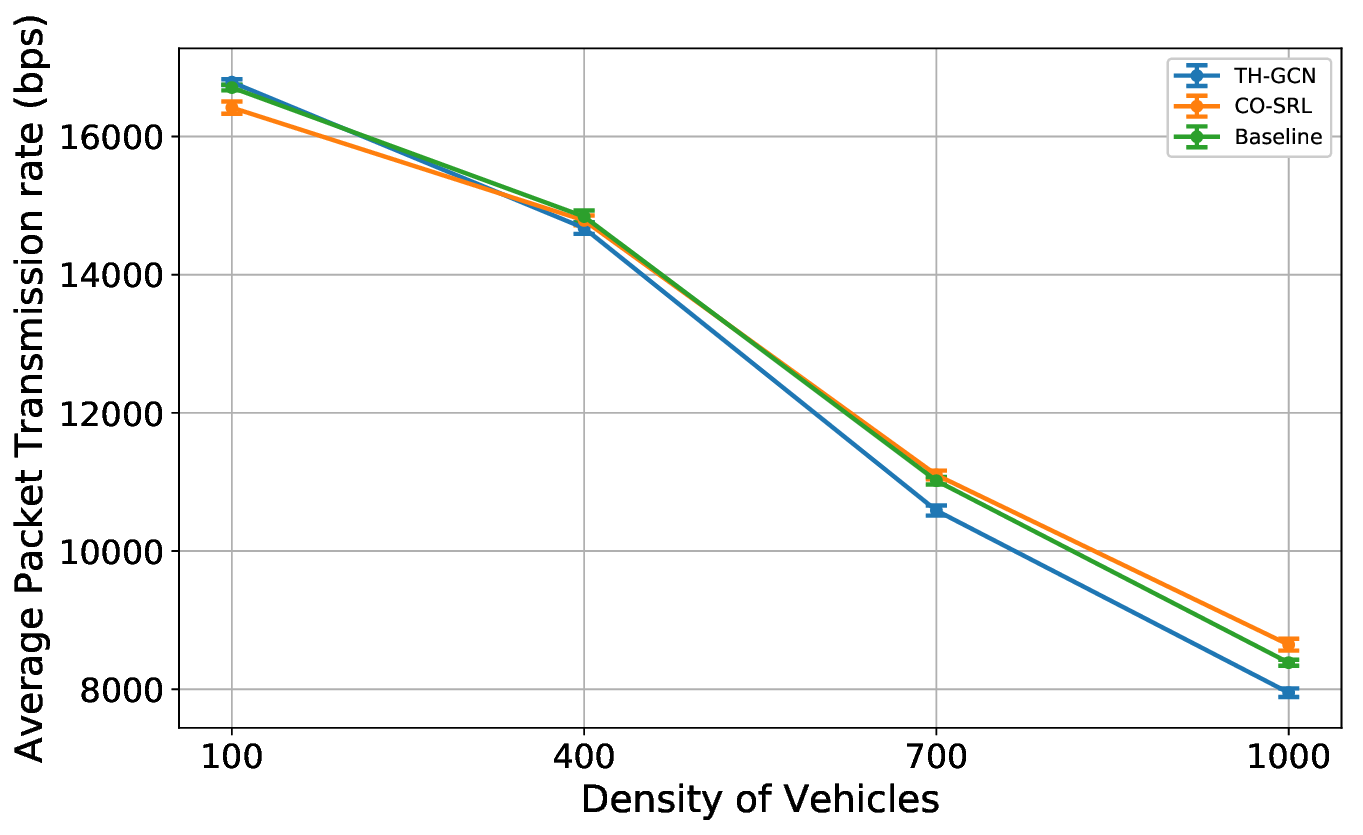}
        \vspace{-4mm}
        \caption{Average Packet Transmission Rate}
        \label{fig:transrate}
    \end{subfigure}
    \;
    \begin{subfigure}{.45\textwidth}
        \centering
        \includegraphics[trim={8pt 5pt 7pt 7pt},clip,width=\textwidth]{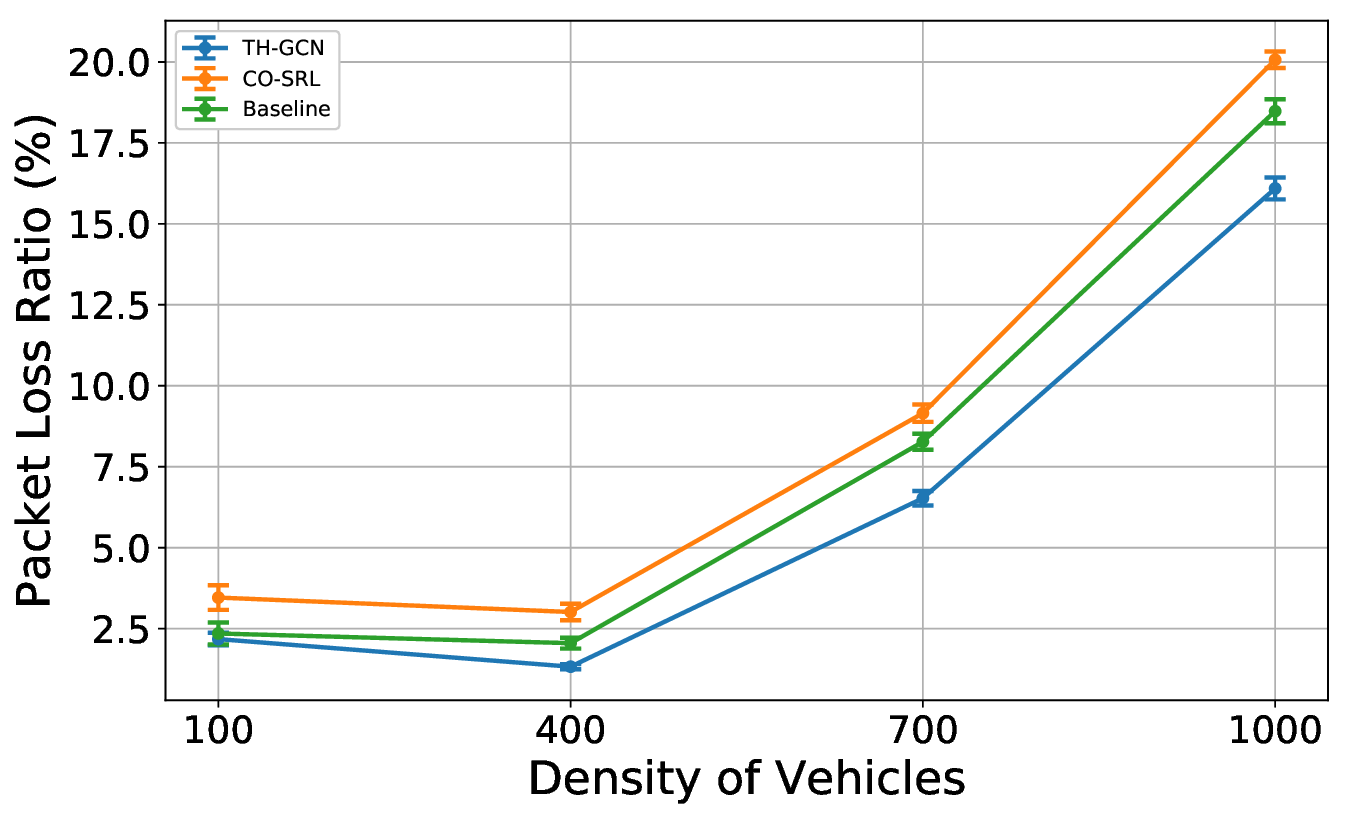}
        \vspace{-4mm}
        \caption{Packet Loss Ratio}
        \label{fig:pktloss}
    \end{subfigure}

    \begin{subfigure}{.45\textwidth}
        \centering
        \includegraphics[trim={8pt 5pt 7pt 7pt},clip,width=\textwidth]{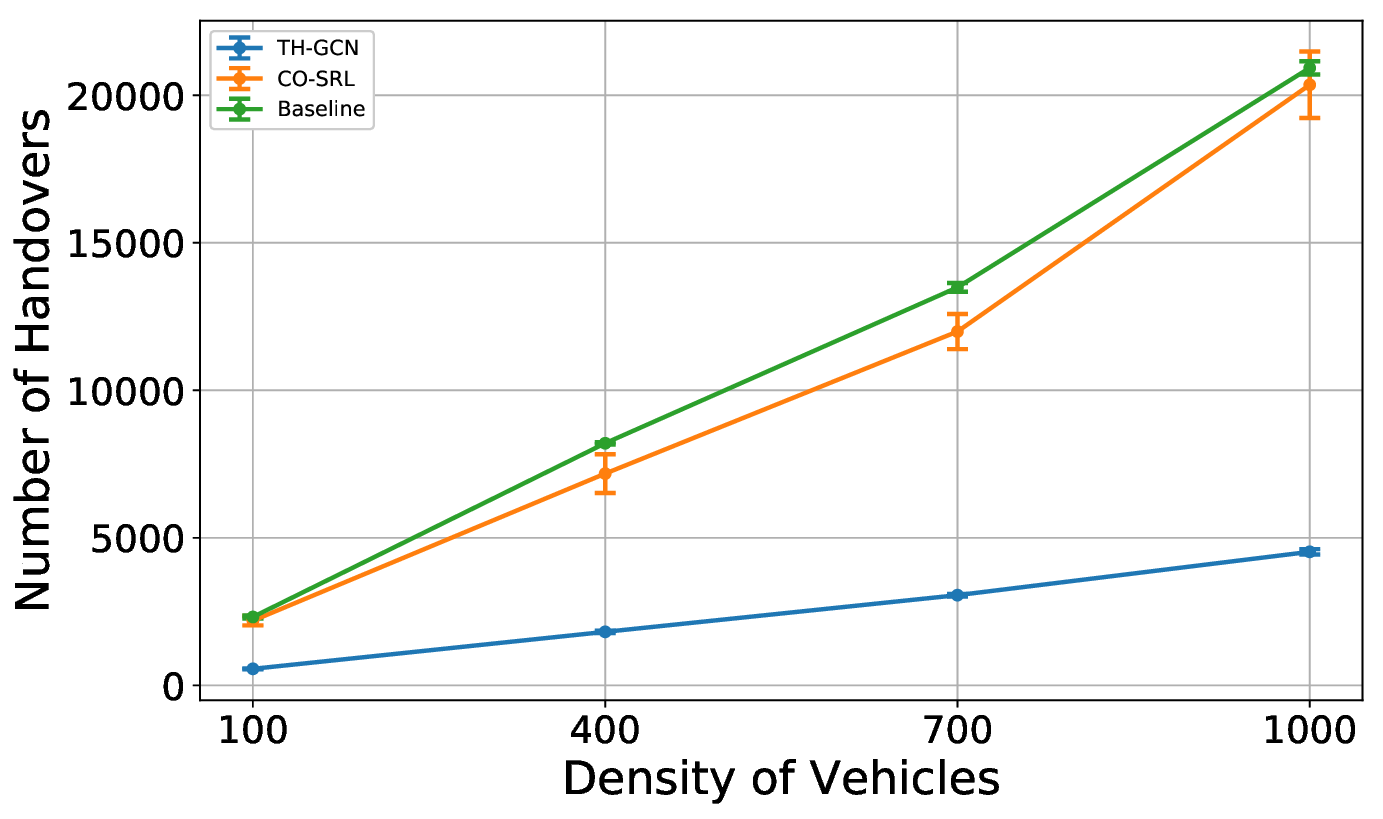}
        \vspace{-4mm}
        \caption{HO Count}
        \label{fig:ho}
    \end{subfigure}
    \;
    \begin{subfigure}{.45\textwidth}
        \centering
        \includegraphics[trim={8pt 5pt 7pt 7pt},clip,width=\textwidth]{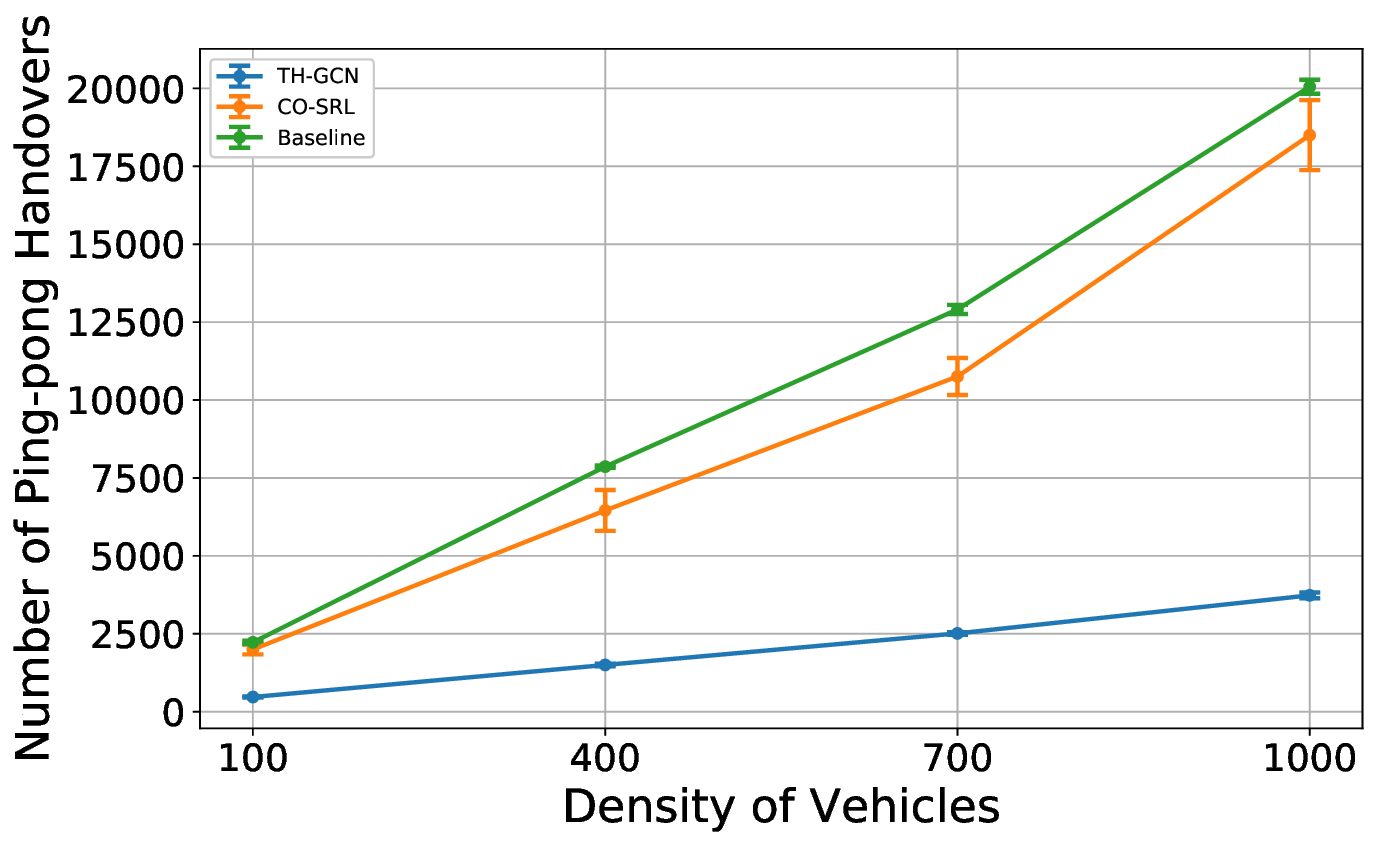}
        \vspace{-4mm}
        \caption{Ping-pong HO Count}
        \label{fig:pho}
    \end{subfigure}

    \caption{Comparison of Evaluation Metrics across various vehicle densities.}
    \label{fig:throughput_cluster}
    \vspace{-2mm}
\end{figure*}

\subsubsection{\textbf{Average SINR}}

TH-GCN consistently achieves the highest SINR across all vehicle densities, with a 10\% improvement over CO-SRL (up to 9.68 dB higher at 100 vehicles), as shown in Figure \ref{fig:sinr}. Though improvements over the baseline are less statistically significant, CO-SRL performs poorly, averaging 26.5 dB. TH-GCN’s superior SINR stems from its graph-based approach and edge-weighted aggregation, enhancing tower selection by considering both local and global conditions. Embedding space optimization through triplet loss further optimizes signal quality across vehicle densities.

\subsubsection{\textbf{Average Simulation Throughput}}
Figure~\ref{fig:throughput} highlights that TH-GCN achieves higher throughput than both CO-SRL and the baseline, particularly under higher vehicle densities, demonstrating its effectiveness in managing network load. This superior throughput stems from TH-GCN’s graph-based design, which provides a holistic view of network states and supports dynamic load balancing across towers. The model's multi-metric consideration through edge features, combined with reduced unnecessary handovers, contributes to sustained high throughput levels even under dense conditions.

\subsubsection{\textbf{Packet Transmission Rate}}
In Figure~\ref{fig:transrate}, while TH-GCN shows slightly lower packet transmission rates at higher densities, it achieved better overall throughput and packet delivery ratio compared to alternatives. The packet transmission rate metric only considers delivered packets, where TH-GCN performs worse than others. This trade-off highlights TH-GCN’s focus on successful delivery over transmission speed. Its efficient resource utilization is valuable in dense scenarios with limited network resources, leading to better service quality despite lower transmission rates.

\subsubsection{\textbf{Packet Loss Ratio and Packet Delivery Ratio}}
TH-GCN consistently achieved the lowest packet loss ratio across all densities, maintaining delivery rates above 97\% and notably outperforming other methods, especially in high-density conditions, as shown in Figure~\ref{fig:pktloss}. This strong performance is attributed to TH-GCN’s advanced tower selection mechanism, which incorporates both current and predicted network states. The edge-weighted aggregation in GCN layers prioritizes stable, low-loss connections, ensuring reliable delivery rates even as vehicle density increases.

\subsubsection{\textbf{Handovers and Ping-pong Handovers}}
TH-GCN demonstrated substantial reductions in the number of both handovers (Figure~\ref{fig:ho}) and ping-pong handovers (Figure~\ref{fig:pho}), supporting more stable connections with fewer interruptions. This improvement is due to the handover decision-making process, where a handover is only triggered if the SINR of the optimal tower selected by the GNN exceeds the current tower’s SINR by a threshold (hysteresis margin). This approach ensures more stable and consistent handover decisions, particularly in high mobility and dense network scenarios.

The analysis demonstrates TH-GCN's robust scalability and superior performance in dense vehicular 5G networks. TH-GCN's integration of both tower-assisted and user-centric views provides a holistic model that excels in delivering stable and reliable handover decisions. By optimizing throughput and minimizing handovers, TH-GCN proves highly effective in dynamic environments, ensuring efficient connectivity while enhancing user experience.

%
%
\section{Conclusion and Future Works}
\label{sec:conclusion_future_works}

This work introduced TH-GCN, a GNN-based model for optimizing handover management in dense 5G vehicular networks. TH-GCN creates a dynamic graph of network conditions by integrating features like signal quality, throughput, and distance. The model’s triplet loss function identifies optimal towers, while a SINR check ensures handovers only occur with significant signal improvement, avoiding unnecessary transitions. Our approach reduces total and ping-pong handovers, improves SINR, packet loss, and throughput, and enhances network performance in high-density scenarios. However, testing in an urban setting with many towers raises concerns about performance in areas with fewer towers. The lack of a hysteresis threshold may cause unnecessary handovers, and reliance on static snapshots limits future condition prediction. Future work will address these issues by adding an adaptive hysteresis threshold and a temporal Graph Convolutional Network to improve predictive accuracy using time-series data.


\bibliography{references}
\bibliographystyle{ieeetr}
\end{document}